\def\year{2022}\relax
\documentclass[letterpaper]{article} 
\usepackage{aaai22}  
\usepackage{times}  
\usepackage{helvet}  
\usepackage{courier}  
\usepackage[hyphens]{url}  
\usepackage{graphicx} 
\usepackage{subfigure}
\def\UrlFont{\rm}  
\usepackage{natbib}  
\usepackage{caption} 
\DeclareCaptionStyle{ruled}{labelfont=normalfont,labelsep=colon,strut=off} 
\usepackage{algorithm}
\usepackage{algorithmic}

\usepackage{newfloat}
\usepackage{listings}
\usepackage{amsmath}

\floatstyle{ruled}
\newfloat{listing}{tb}{lst}{}
\floatname{listing}{Listing}
\title{Neural combinatorial optimization beyond the TSP:\\Existing architectures under-represent graph structure}

\author {
    Matteo Boffa,\textsuperscript{\rm 1 \rm 2}
    Zied Ben Houidi, \textsuperscript{\rm 1}
    Jonatan Krolikowski \textsuperscript{\rm 1}
    Dario Rossi \textsuperscript{\rm 1}
}
\affiliations {
    \textsuperscript{\rm 1} Huawei Technologies France SASU \\
    \textsuperscript{\rm 2} Politecnico di Torino 
}

\usepackage{bibentry}

\usepackage{color}

\begin{document}

\maketitle

\begin{abstract}
Recent years have witnessed the promise that reinforcement learning, coupled with Graph Neural Network (GNN) architectures, could learn to solve hard combinatorial optimization problems: given raw input data and an evaluator to guide the process, the idea is to automatically learn a  policy able to return feasible and high-quality outputs.
Recent work have shown promising results but the latter were mainly evaluated on the travelling salesman problem (TSP) and similar abstract variants such as Split Delivery Vehicle Routing Problem (SDVRP). In this paper, we analyze how and whether recent neural architectures can be applied to graph problems of practical importance. 
We thus set out to systematically ``transfer'' these architectures to the Power and Channel Allocation Problem (PCAP), which has practical relevance for, e.g., radio resource allocation in wireless networks. 
Our experimental results suggest that existing architectures (i) are still incapable of capturing graph structural features and (ii) are not suitable for problems where the actions on the graph change the graph attributes. On a positive note, we show that augmenting the structural representation of problems with Distance Encoding is a promising step towards the still-ambitious goal of learning multi-purpose autonomous solvers. 
\end{abstract}


\section{Introduction}
Graphs are well-studied objects used to represent a plethora of practical combinatorial problems. Solving such graph-related combinatorial problems represents an everyday challenge for vital tasks such as logistics, resource allocation or scheduling. Many of these problems are NP-hard and 
optimal solutions are attainable only for trivial graph size, so that in practice problem relaxation, or  approximation algorithms or various search heuristics are used.

Recent years have seen the rise of deep learning models aimed at  replacing such handcrafted  heuristics, by leveraging advances in two areas of machine learning. First, graph representation learning extracts rich features from raw input graph data. Second, deep reinforcement learning allows agents to ``navigate'' in a search space and find an optimal strategy to maximize a given objective. Most existing models employ jointly these two strategies and train agents that (i) take a graph with node and edge features as input, (ii) encode it, often using a graph neural network (iii) sequentially obtain an action for each node/edge using a decoder and (iv) guide the agent with a simple environment evaluating the agent's strategy.  
However, looking more closely at the literature applying  graph neural network (GNNs) and reinforcement learning to practical combinatorial problems, two observations arise. Existing solutions are either (i) admittedly tailored to specific problem instances (e.g. extensively using expert-based feature engineering or neural network design~\cite{9472828}, or not transferring to graphs of different sizes~\cite{DRL:allocation}) or are (ii) generic by design, but  limitedly evaluated on the very same problem, namely the Travelling salesman problem (TSP) and variants thereof~\cite{bello2016neural}. 

Although we agree it is useful to compare architectural solutions on the same ground, we argue that this may lead to over-fitting the design to a single problem. In practice, it is today unclear whether all the recent advances applied and evaluated on the TSP~\cite{NIPS2015_5866,bello2016neural, Dai2017, inbook, kool2018attention, ma2019combinatorial, nowak2017revised}, are useful for other graph combinatorial problems of practical interest. We thus set out to first systematically ``transfer'' these architectures from the TSP to other problems with high practical relevance, related to radio resource allocation in wireless networks. There, the input is a graph representing the \emph{access points} (APs) and the user \emph{stations} (STAs) connected to them, the output is the allocation of frequency channels and power levels to each AP. The goal is to achieve both low interference and good signal quality.

When adapting existing ``out of the box'' architectures, we observe subtle yet crucial differences between TSP (and similar variants) and the practical problem we consider (Sec.~\ref{problem_transfer_sec}). First, unlike the Euclidean TSP graph, where node features encode graph structure in the form of \emph{Euclidean coordinates of each node}, our input graph contains mainly \emph{distances between nodes} (in the form of radio path losses). Second, while in TSP, algorithmic actions have no impact on the graph structure and attributes (cities will not move), in radio resource allocation, assigning frequencies or power levels to certain nodes affects the attributes of other nodes (such as signal quality or interference). As we shall see, 
these two differences are the root of poor performance exhibited by state of the art architectures, questioning their generality and relevance beyond the TSP.

The rest of this paper is organized as follows. Sec.~\ref{sec:related} overviews related work. Sec.~\ref{problem_transfer_sec} adapts existing solutions to our problem. Evaluations in Sec.~\ref{pars_destruens} confirm our two observations, and  Sec.~\ref{sec:rescue} points to distance encoding as a promising direction to overcome limitations of current architectures.

 
\section{Related work}\label{sec:related}



The foundations for using deep learning to solve combinatorial problems are Pointer Networks~\cite{NIPS2015_5866}
which enhance pre-existing sequence-to-sequence approaches, enabling them to learn solutions to the TSP on variable-sized instances.
A pointer network is an encoder-decoder Recurrent Neural Network (RNN) in which the problem inputs are converted into \emph{codes} by an encoder, before being fed to the following generative model or decoder. The latter applies a content-based attention using the hidden-state of each decoding block to point back at each node in the encoded input. When applied on a combinatorial problem like the TSP, the attention mechanism helps the neural network to generate pointers towards the input sequence, thus providing the order in which the nodes are visited.
Since this seminal work, almost all of the follow-up architectures adopted a similar philosophy with an encoder-decoder design where the encoder learns representations of the input graph, and the decoder uses those representations to sequentially generate an action for each node or edge. 

Next, while the first pointer networks \cite{NIPS2015_5866} learn in a supervised manner (using optimal solutions as labels), \cite{bello2016neural} later leverage a similar model but adopt Reinforcement Learning instead. This design choice is 
sensible since optimal solutions are often not available for large enough instances and since it is often easier to define a reward or regret that reflects  solution quality. For example, in the case of the TSP, the tour length can be used as a regret. Many others have used reinforcement learning since then. A notable example is the work of \cite{NIPS2018_8190}, which  represents an important milestone for Deep Learning applications to graph combinatorial problems: firstly, following the line of \cite{44871}, the authors observe that the problems of interest do not necessarily need sequential order. This means that the so-far used RNN-encoder-decoder models might actually be unnecessary and could easily be substituted by less computationally demanding architectures, perhaps even simple linear projections. Secondly, facing the Vehicle Routing Problem (VRP), a problem similar but slightly more complex than the TSP, they are the first to deal with what we call in this paper the distinction between dynamic \emph{action-dependent} and \emph{static} node features, the former of which previous architectures could not deal with. As we will see in Sec.~\ref{problem_transfer_sec}, this issue is highly important in the context of our work.
After an unsuccessful encoding attempt using  Graph Convolutional Networks (GCNs) proposed by \cite{DBLP:journals/corr/KipfW16}, \cite{NIPS2018_8190} opted to only encode node-specific information (i.e. coordinates and demand values), with a 1-dimensional convolution layer, and did not include any adjacency information. 
%

Later, building on the work of \cite{NIPS2017_7181} and \cite{velickovic2018graph}, \cite{kool2018attention} were able to tackle problems like the TSP and the VRP using an attention-based GNN model as encoder, namely the \textit{Graph Attention Network} (GAT). This, coupled with a sequential decoder that keeps track of the current context (e.g. which nodes have been visited so far), greatly improved results. 
As already mentioned, Graph Convolutional Networks are an alternative to GATs for problem encoding, leveraged by \cite{joshi2019efficient} when tackling the TSP. Similarly to the GAT model chosen by \cite{kool2018attention}, nodes become aware of their surroundings through a message passing procedure. However, the concept of distance can be applied directly as edge weights here, while the GAT needs to learn these features through its \emph{attention} mechanism. 

Very recently, \cite{joshi2020learning} proposed to create a unified benchmark to systematically test the generalization capability of several existing design choices. Focusing on the importance of transferring models trained on smaller instances to larger ones, the authors check whether and how much past design choices 
favour such transfer.
Their systematic work pinpointed some limitations and stressed the importance of taking generalization into account, both during design and evaluation.
In this paper, we extend their work to another dimension: while they attempted to scale ``vertically" to networks of larger sizes, we want to assess whether the latest neural network design choices are of interest to practical problems that are structurally
different from the TSP. 
Given a topology with some node and edge features and actions to apply to nodes or edges, how difficult it is to adopt current architectures? And how well would they perform compared to a well-thought heuristic? 

Answering these questions, we conclude that existing architectures are inadequate to our use case. As per the recent studies of \cite{gilmer2017neural}, \cite{chen2019utilizing} and \cite{10.1145/3452296.3472902}, we found that all existing models, even if they are efficiently handling node information for tasks such as node classification \cite{you2018graph} or node property prediction \cite{DBLP:journals/corr/KipfW16}, fail to capture edge features well. Following recent works, e.g.\ \cite{li2020distance} and  \cite{yin2020revisiting} that point out that current GNN models suffer from  similar shortcomings, we implement their proposed novel approach \emph{distance encoding} and evaluate it.

Finally, it is worth mentioning that our work, and related work accordingly, focus on learning an end-to-end solution or heuristic to the problem. We refer the reader to the recent survey of \cite{cappart2021combinatorial} for a broader perspective about learning for combinatorial optimization in general.

\section{Problem transfer}\label{problem_transfer_sec}
In this section, we qualitatively describe the process of adapting existing architectures
from well-studied problems to our practical radio resource allocation one. We first describe the problems (Sec.~\ref{use_cases}) and  overview existing architectures (Sec.~\ref{sec:arch:overview}). We then highlight novelties introduced by our problem (Sec.~\ref{sec:problem:diff}) and conclude with some implementation and evaluation considerations (Sec.~\ref{sec:implementation}).

\subsection{Use cases}\label{use_cases}
We now present three combinatorial problems relevant for this paper: the \textit{Traveling Salesman Problem} (TSP), the \textit{Split Delivery Vehicle Routing Problem} (SDVRP), and our \textit{Power and Channel Allocation Problem} (PCAP). While the neural combinatorial optimization literature has limitedly studied the first two, we here contrast the first and the third.
\subsubsection{TSP}
The well known TSP~\cite{christofides1972bounds} asks for the shortest tour a salesman can take, visiting each of a given set of cities once before returning home. More formally, the objective is to find the shortest \emph{Hamiltonian} cycle  
in a graph 
with given edge lengths.
In the Euclidean TSP,  the edge lengths are the Euclidean distances between the nodes. 
Existing architectures trained with reinforcement learning on this problem use the tour length as regret.

\subsubsection{SDVRP}
The SDVRP is a generalization of the Capacitated Vehicle Routing Problem (CVRP)~\cite{toth2014vehicle}. There, a number of delivery vehicles depart from a designated \textit{depot}, visit all  \emph{customers}  and then return to the depot. 
Each customer has a specific demand, and each vehicle a capacity. The sum of customer demands are not allowed to exceed the assigned vehicle's capacity. The objective is to serve all customers while minimizing the sum of the cycle lengths, defined again as Euclidean distances.
Note that CVRP with one vehicle and infinite capacity is equivalent to the TSP. In the SDVRP, the nodes can be visited by multiple vehicles, each satisfying  a part of the demand. 
Unlike the TSP, for which we perform empirical analysis, we present the SDVRP mainly for illustrative purposes as it represents an interesting property which we discuss later in Sec.~\ref{sec:problem:diff}.

\subsubsection{PCAP} While the two previously mentioned problem classes are well-known abstractions of real world applications, the PCAP is directly derived from the radio resource management (RRM) of IEEE 802.11 Wireless Local Area Networks (WLANs). In large-scale WLANs consisting of fleets of APs, choosing adequate levels of AP channels, bandwidths and transmission powers is a vital task to manage scarce radio resources. Broadly speaking, WLAN RRM has three objectives: (i) optimal coverage for the users STAs served by the network; (ii) minimizing interference among both APs and STAs; (iii) load-balancing of associated STAs among APs. The PCAP is challenging mainly due to the fact that power, channel and bandwidth allocation are discrete configurations, making PCAP a combinatorial optimization problem. In fact, the channel allocation problem alone is NP-complete~\cite{surveyCAP}.

The inputs to solve PCAP are, pathloss measurements among APs/STAs, available channels and power levels, and STA demands, whereas coordinates of APs and STAs are generally not available.
The underlying structure of PCAP is a graph, in which transceivers (APs and STAs) represent the nodes, connected via edges representing channel interference, where the weight of each edge depends on radio conditions (pathloss, antenna gains, etc.). 
A reward metric evaluating the quality of any allocation can be then computed based on STA coverage, interference, and AP loads. 
Given that PCAP is a graph-based combinatorial optimization problem for which the objective value of a solution can be evaluated efficiently, it is a good  use case for application of Deep Learning architectures originally designed for TSP.





\subsection{Overview of existing architectures}\label{sec:arch:overview}
Our goal in this work is to systematically assess to what extent existing neural architectures, evaluated on abstract problems such as TSP and VRP, are applicable to other problems such as PCAP. We now provide a schematic overview of existing design choices used in this paper.

\begin{figure*}[htb]
    \centering
    \subfigure[TSP adaptation]{\label{TSP_adapt}\includegraphics[width=.24\linewidth]{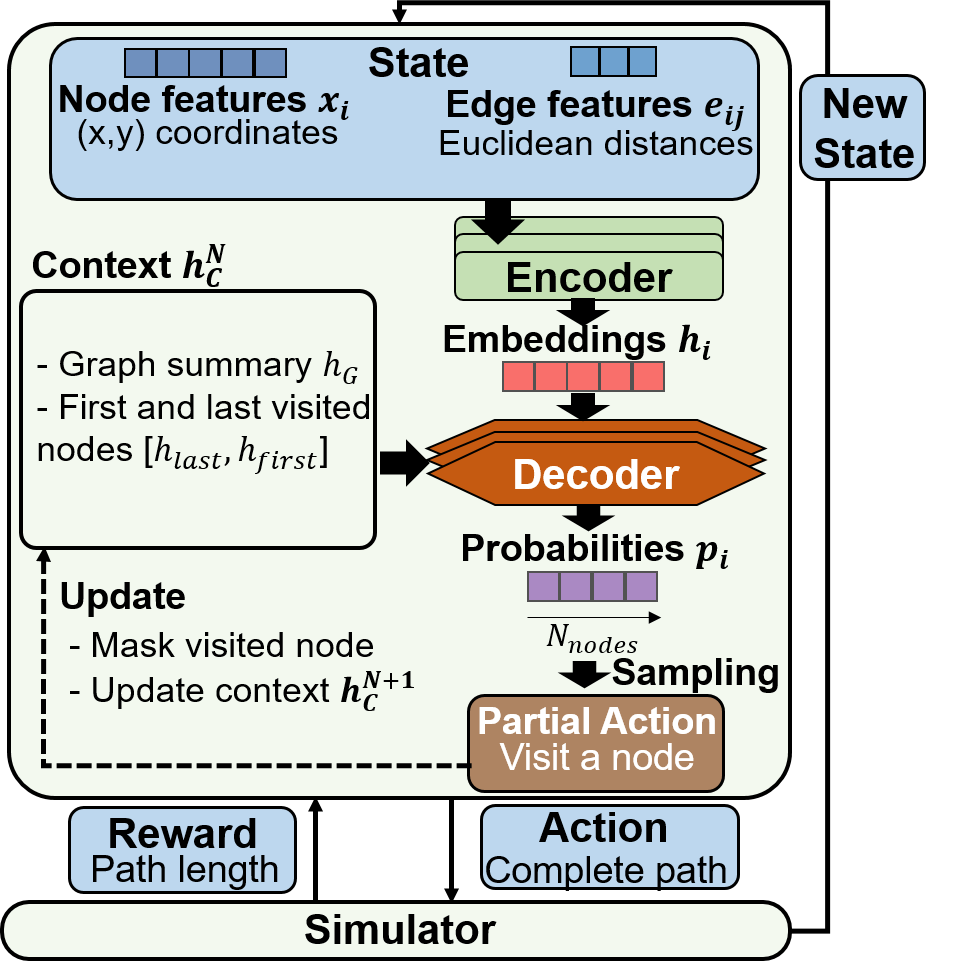}}
    \hspace{0.5cm}
    \subfigure[SDVRP adaptation]{\label{SDVRP_adapt}\includegraphics[width=.24\linewidth]{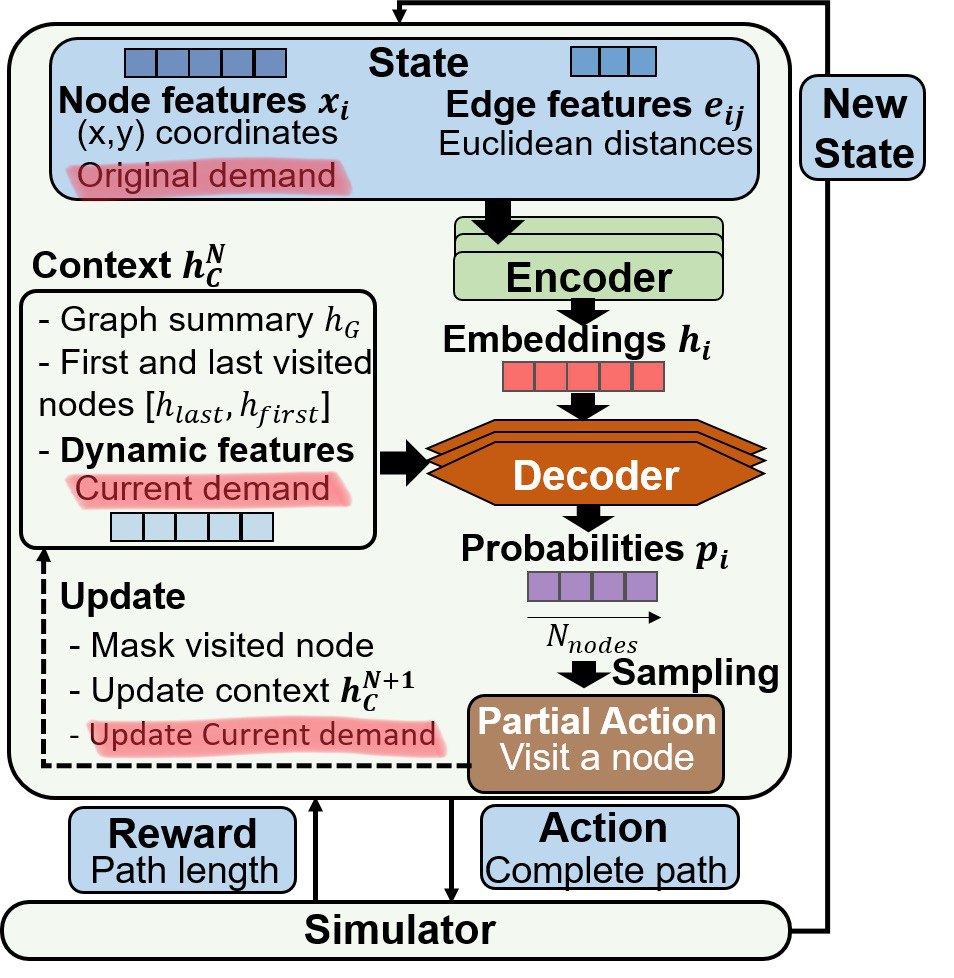}}
    \subfigure[PCAP adaptation]{\label{PCAP_adapt}\includegraphics[width=.24\linewidth]{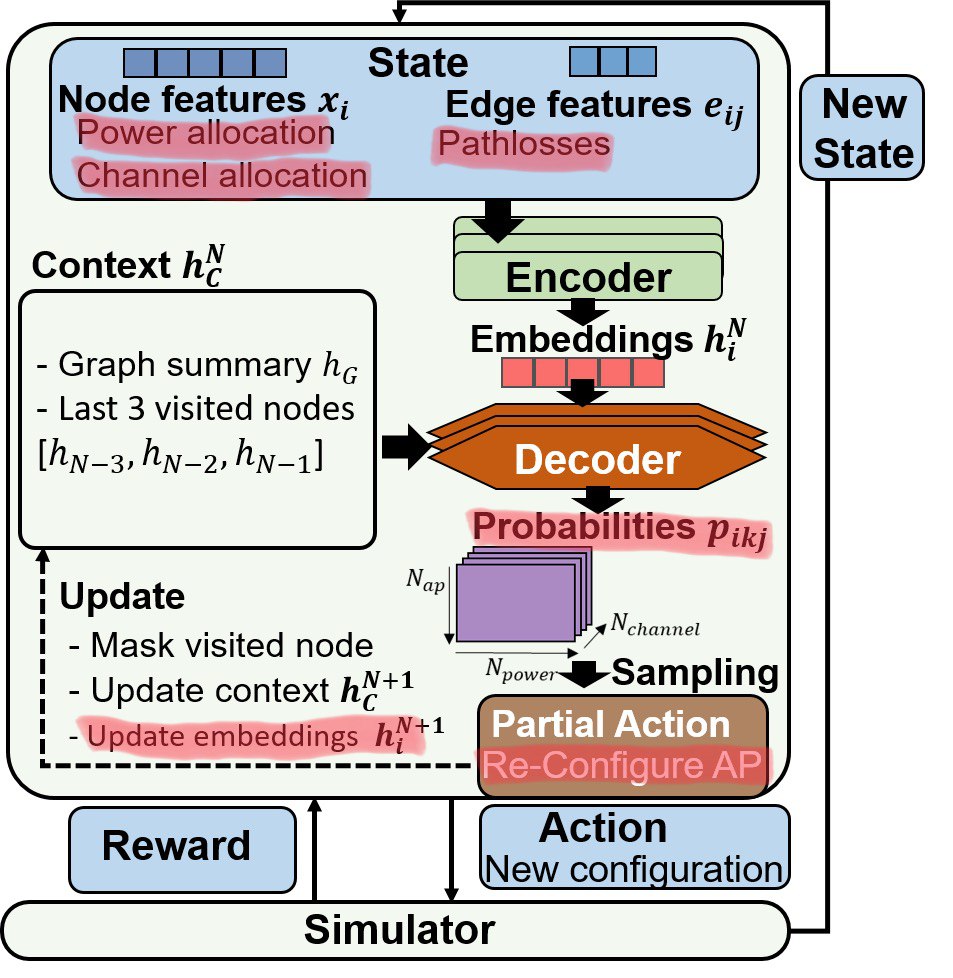}}
    \caption{Abstraction of the generic pipeline used in recent literature. Red highlights indicate changes from the TSP.}
    \label{enc_dec_pipeline_1}
\end{figure*}

As mentioned earlier, state-of-the-art solutions rely on \textit{encoder-decoder} architectures, where the encoder learns a representation of the input problem and the decoder uses that to output a solution.
~\cite{joshi2020learning} provide a modular implementation and useful abstractions to synthetically summarize existing approaches for the TSP. Building on their work, we show the generic pipeline shared by many such recent work in Fig.~\ref{enc_dec_pipeline_1}. We first show in Fig.~\ref{TSP_adapt} the pipeline for the TSP problem. We then mirror it for the SDVRP (Fig.~\ref{SDVRP_adapt}) and our PCAP problem (Fig.~\ref{PCAP_adapt}).

The encoder, a GNN, first receives as input problem instances or graphs of various sizes in the shape of node and edge features. Node features are, for example, coordinates in the TSP case, while edge features comprise the adjacency matrix of the underlying graph. 
The GNN learns  fixed-size representations (\emph{embeddings}) which mirror each input and its  surroundings.
\cite{kool2018attention} propose to leverage the aforementioned GAT, while ~\cite{joshi2019efficient} later introduced the GCN encoder. Finally, ~\cite{joshi2020learning} include the two options in their common modular benchmark and demonstrate that the latter is capable of obtaining more faithful representations. In our work, we evaluate both options. However, unless otherwise mentioned, the GNN of~\cite{joshi2019efficient} is our default choice for the encoder; we refer to it as \emph{TSP state-of-the-art} in later sections.

Next, the graph and node embeddings are passed to the decoder. A popular~\cite{Dai2017,kool2018attention,ma2019combinatorial,inbook,bello2016neural,NIPS2015_5866} design choice is to use a decoding loop that sequentially picks an action for each node, until a stopping condition (i.e. all nodes visited) is reached. This means that the decoder needs to handle partial solutions. Conversely, the research of \cite{joshi2019efficient} adopted a one-shot decoding strategy in which the entire solution (path) is computed once and for all nodes, with no  intermediate decisions. However, since~\cite{joshi2020learning} include both options in their modular framework and prove that the sequential option
has better performance, we adopt the sequential attention-based decoder used by~\cite{kool2018attention}. 

Solving the TSP, the decoder is run iteratively, generating at each step a learned probability distribution over all possible next nodes. Sampling from this probability distribution picks the next node to visit, appending it to the current \emph{partial solution}. The selected node is then masked so that it is not visited in the next iterations of the decoder. This way,  the decoder outputs a probability over only the remaining nodes each time. Note that at each step, the decoder receives also an updated context vector describing the current state of the partial solution: in the case of TSP, it contains the embeddings of the first and last visited nodes.

A very similar architecture has been used to  solve other problems such as the SDVRP, the Orienteering Problem and the Prize Collecting TSP~\cite{kool2018attention, NIPS2018_8190}. In Fig.~\ref{SDVRP_adapt} we illustrate how the same architecture is adapted for the SDVRP problem. Compared to TSP, node features are augmented to account for demands. This also has consequences on the context as we will see later. Finally, as illustrated in Fig.~\ref{PCAP_adapt}, we adapt the same architecture for the PCAP, first with as few changes as possible: the node features are the power levels and channels, and edge features contain path losses between the various transceivers. The action is taken as well per node, and the decoder learns to generate a probability distribution of shape $\#\text{APs} \times \#\text{power levels} \times \#\text{channels}$, that is the probabilities of assigning power levels and channels to each AP. One single action is then taken at each step in the decoder loop by sampling from the learned distribution. The chosen AP is masked for the next decoder iteration, such that  only the remaining available options remain.
More generally,  the decoder loop may also iterate over edges as well in case that actions are edge specific rather than node specific.



\subsection{Differences between the problems}\label{sec:problem:diff}
The three use cases mentioned in Sec.~\ref{use_cases} are combinatorial optimization problems that can be represented with graphs, i.e.\ a set of node and edge features. The respective actions are all related to nodes\footnote{In other combinatorial optimization problems such as the network design problem, the actions affect foremost the edges.}. However, looking more closely, there are differences that affect the ``transfer'' of architectures from one problem to another.

\subsubsection{Node vs. Edge features}\label{node_edge_sec}
Generally, a graph $G=(V,E)$ is characterized by the \emph{node set} $V$ and the \textit{edge set} $E$, together with their respective features. The former represent entities and their characteristics: cities with their coordinates in TSP, customers with their locations and demands in SDVRP, and APs with e.g.\ antenna gains and configurations as well as STAs with e.g.\ their demands for PCAP. 
The latter encode relationships between these entities, in terms of their mere existence (e.g.\ possibility to travel from city A to city B in TSP), the distance (e.g.\ between two cities), or various characteristics of links (e.g.\ shadowing between transceivers). 

In some real-life examples, there is a direct dependency between node and edge features: for instance, if we provide geographic coordinates, we can easily compute the straight-line distances. In other cases, such conversion can only be approximated: for example, given the geolocations of two APs, the pathloss between them is subject to different types of attenuation not obvious from their positions only. Finally,  some conversions, are highly complex or impossible: it is very difficult to determine a STA's meter-exact geolocation from pathloss data. This fact leads us to the first major difference between the TSP/SDVRP and PCAP: unlike the TSP where node features (coordinates) convey also information about edge features (distances between nodes), the ``distances'' between nodes in the PCAP is contained only in the edge features in the form of path losses.  
This puts stress on encoders that \textit{must capture well both node and edge features}.

\subsubsection{Static vs. action-dependent graph features}\label{static_vs_dynamic}
As discussed in Sec.~\ref{sec:arch:overview}, during each iteration of the decoder loop, the agent increments the current partial solution with the new action and modifies the context for the next iteration accordingly.
For TSP, this means masking already visited nodes, and tracking the two end nodes of the current partial path. 
For SDVRP, the residual demands need to be included ``manually''.
The picture becomes even more complex for PCAP, since a partial action has consequences on the current STA coverage, AP loads, and interference between nodes. 

Adopting a sequential decoder, we identify different types of problems that depend on the consequences of partial solutions on the features of yet unassigned nodes: (i) problems where incremental actions do not change input features: this is the TSP where the partial solution does not impact the coordinates of the yet unvisited nodes. We speak of \emph{static features only}.  (ii) Intermediate actions change some input features while others remain unaffected. This is the case of SDVRP where the residual demand is affected, but the geo-coordinates (and consequently distances) are not. We speak of a problem that has both \emph{static and dynamic features}. 
(iii) The partial solution  affects all node features. This is the case for PCAP where we have \emph{dynamic features only}.

\begin{figure}[htb]
    \centering
    \subfigure[TSP]{\label{TSP_decoder}\includegraphics[width=.47\linewidth, height=.45\linewidth]{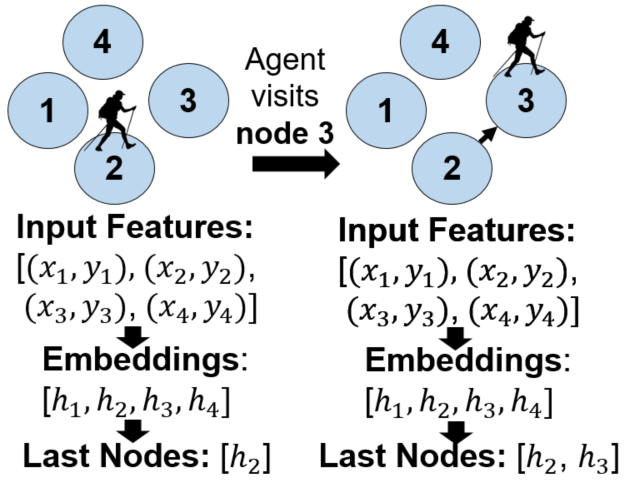}}
    \hspace{0.1cm}
    \subfigure[PCAP]{\label{PCAP_decoder}\includegraphics[width=.47\linewidth, height=.45\linewidth]{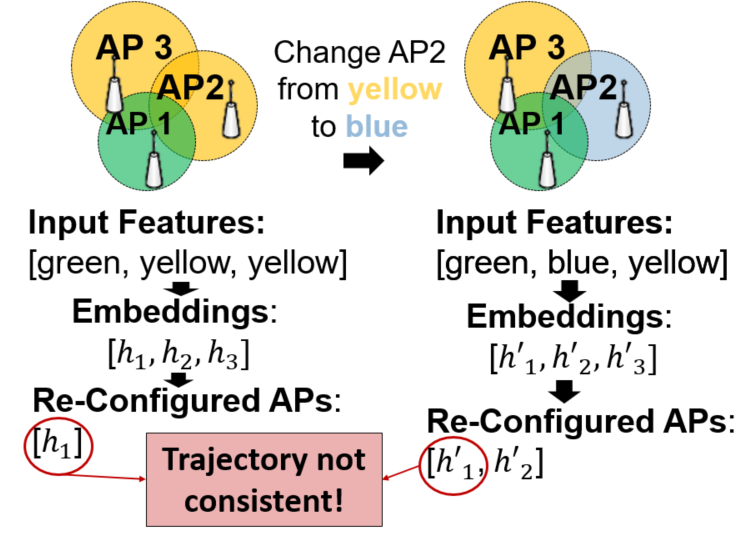}}
    \caption{Impact of agent's intermediate decisions during decoding loop: static and action-dependent features}
    \label{decoder_issue}
\end{figure}

As illustrated in Fig~\ref{TSP_decoder}, for ``static features only'' and  ``static and dynamic features'', we can operate as follows: First, obtain fixed embeddings $h^L_i$ of the problem, which are immutable and independent from the agent's actions. 
Then, during decoding, simply include in the decoding context the representation of (the consequences of) the current partial solution (e.g. the residual demands for the SDVRP or the last node visited in the TSP).
This is not possible for our problem where keeping track of the current context becomes a challenge for two reasons: (i) When the agent picks an incremental action in iteration $t$, it also modifies the  graph characteristics (e.g. changes the initial power level from ``low'' to ``high''); thus, at the next decision at time $t+1$, the agent should not rely anymore on the ``old embeddings'', since they might be not consistent with the current state of the graph. Even (ii) keeping track of the recent ``trajectory'' (e.g. last configured APs) becomes hard, since, as shown in Fig~\ref{PCAP_decoder}: once an AP is modified, due to the GNN message passing, the entire set of the embeddings change and the ``new history'' is no longer coherent with the past one. 

To counter problem (i), in our adaptation, we implement a re-embedding procedure: every time an action is taken, the encoder computes again the updated embeddings before proceeding with the next decoding iteration. However, re-embedding does not solve (ii).  In the absence of remedies, we track the last three configured APs as decoder context and test this issue empirically. 


\subsection{Implementation and evaluation details}\label{sec:implementation}
 We build on the modular implementation released by \cite{joshi2020learning}\footnote{\url{https://github.com/chaitjo/learning-tsp}}. It consists of a framework to train TSP instances of various sizes using different encoders (e.g. GAT vs. GGCN) activating and deactivating on-demand certain features (e.g. with vs. without decoder), using various hyperparameters etc. Our new study extends their work as we adapt the encoder and decoder inputs and outputs to the PCAP, and then we train the resulting model.
 

Our learning agent learns with the help of a custom WLAN simulator that is able to compute STA coverage, interference and AP load based on channel and power configuration (actions). As our reward for the learner we use a variant of the objective function defined by \cite{ahmed2006successive}. It is proportional to coverage, inversely proportional to interference, and inversely proportional to the AP load.

We tested on scenarios with varying numbers of APs, ranging from 9 to 32 APs. Whenever possible, exact results were obtained by complete enumeration. Since this is extremely costly in terms of time resources, it was feasible only for the network sizes of 9, 14 an 16 APs. This is why a \emph{local search} heuristic was employed to find  solutions for larger instances. Roughly, the heuristic  outputs a locally optimal power configuration with respect to the reward function.  Given an initial configuration, it varies the power level for each AP separately within a specified range, while keeping the remaining configuration constant. The best individual power level for each AP, together with its overall reward gain, are recorded. After an iteration over all APs, the best improvement is compared with a simultaneous change of all APs into their individual directions, and the process is repeated for the best configuration found. If no further improvement can be achieved, the algorithm terminates.

The metrics we used to validate the results are: (i) the \emph{optimality gap} whenever we could exhaustively compute the optimal solution, or the \emph{heuristic gap} which is the gap to the best solution found by our local-search based heuristic. The gap is described as $gap_i=\frac{r(x_{i}^{\ast})-r(\hat{x}_i)}{r(x_{i}^{\ast})}$, where $\hat{x}_i$ is the model's solution for batch $i$ and $x_{i}^{\ast}$ the exact/heuristic one, and $r$ is the reward; 
(ii) \emph{reward difference}, which measures the difference between the rewards of the two competing solutions when applied on the same graph instance. An interval of confidence around the mean is also computed to ensure the difference is significant.

APs and STAs were randomly distributed in the plane  for each scenario: this resulted in 32 different instances for each scenario. We  used $50$ different seeds for training and validation in order to ensure robustness of results: $50 \times 32=1600$ attempts were evaluated per scenario so that the results in Fig.~\ref{distance_encoding_pap} alone represent about 300 GPU-hours.


\section{Transfer results} \label{pars_destruens}


We first evaluate whether the current state-of-the-art architectures, valid for the TSP, perform well on PCAP. Firstly, on Sec.~\ref{transfer_as_it_is_sec}, we evaluate the existing architectures introducing ``as few changes as possible'', only adapting the inputs and outputs of both the encoder and the decoder. Then, in Sec.~\ref{decoder_issue_sec} and Sec.~\ref{encoder_issue_sec}, we empirically test to what extent the overall behaviour is influenced by the problem differences highlighted in Sec.~\ref{sec:problem:diff}.

\subsection{Transfer ``as it is''}
\label{transfer_as_it_is_sec}

\begin{figure}[t]
\centering
\includegraphics[width=.9\linewidth]{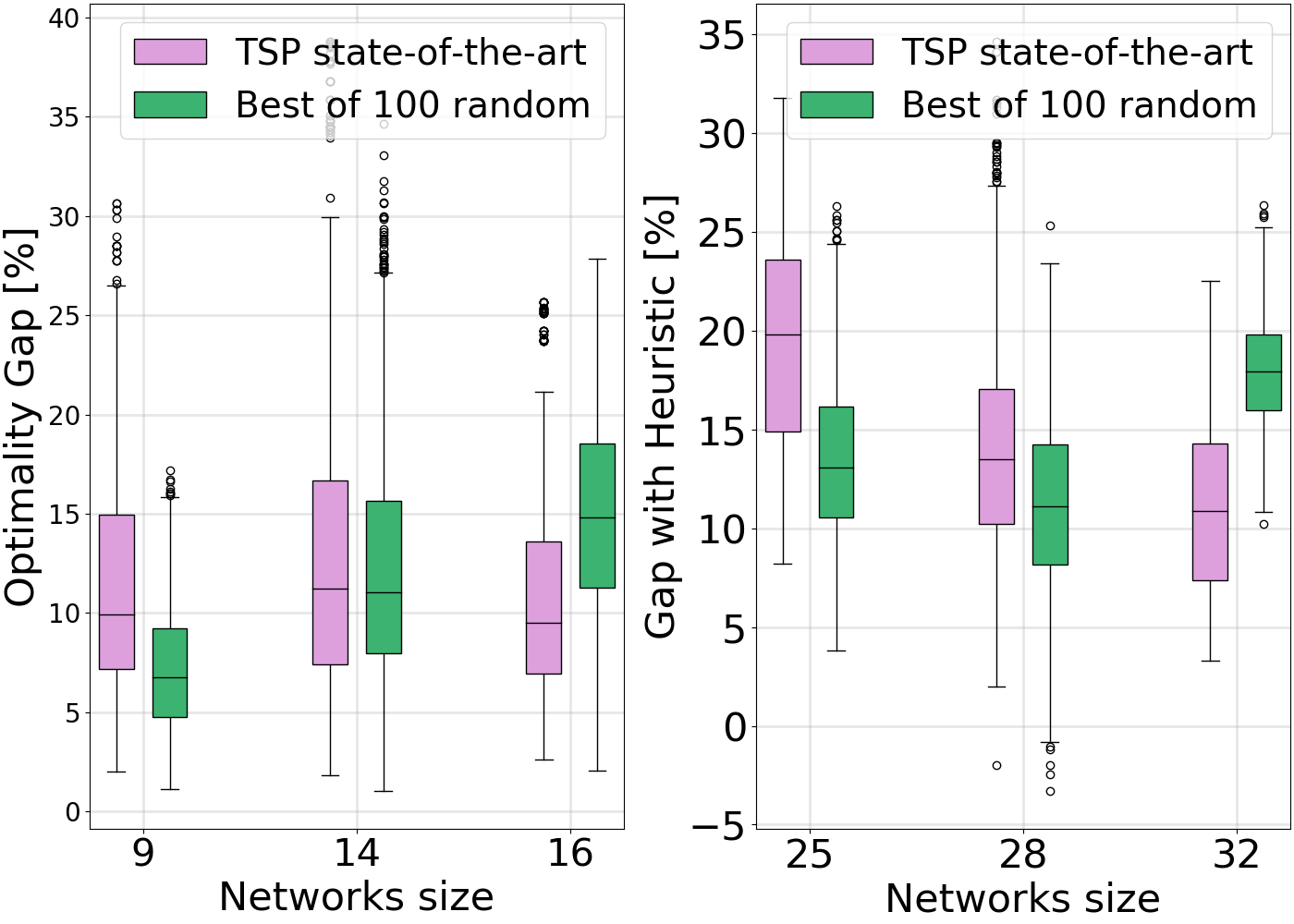}
\caption{Best TSP architecture vs Random Decision Maker}
\label{transfer_minimum_fig}
\end{figure}

Fig~\ref{transfer_minimum_fig} shows the optimality gaps, and gaps from heuristic, achieved by the state-of-the-art model, compared to a random decision maker. Each approach is given 100 attempts to find the best solution for each scenario, and  only the best guess is ultimately kept.
As can be seen, the best random guesses outperform those of the TSP solver in many scenarios: otherwise stated, \emph{we gather that current state-of-the-art choices which are valid for the TSP, are not necessarily transferable to other combinatorial graph problems}.
To shed further light on the reasons behind this poor performance, and having in mind the differences we highlighted in Sec.~\ref{sec:problem:diff}, in the remainder of this section, we drill down further on  the root causes of the above observation.

\subsection{The encoder under-represents structure} \label{encoder_issue_sec}

\begin{figure}[t]
\centering
\includegraphics[width=.9\linewidth]{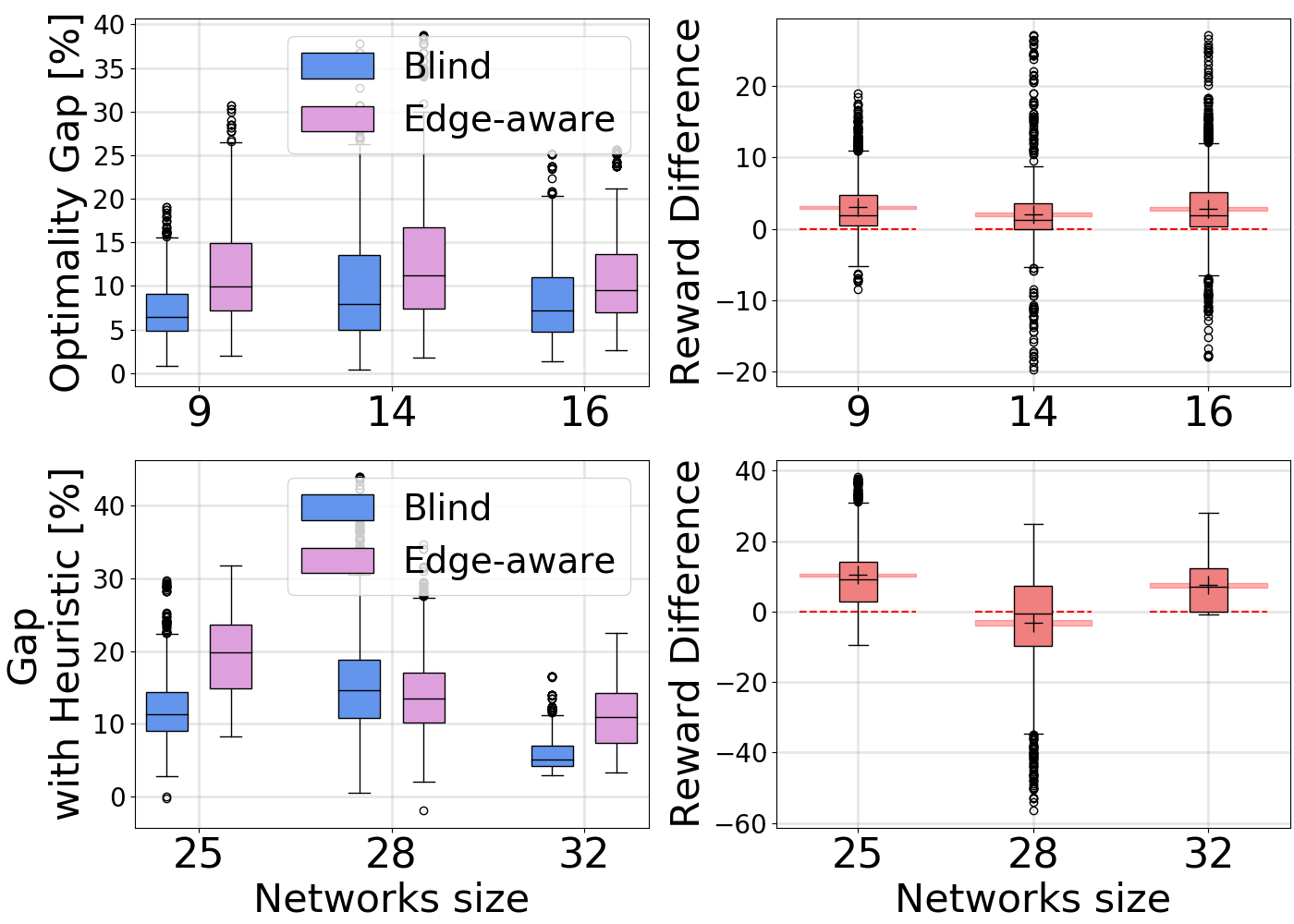}
\caption{Blind vs Edge-aware models}
\label{encoder_issues_fig}
\end{figure}

As pinpointed in Sec.~\ref{node_edge_sec}, edge features are highly important for PCAP while they can be derived from node features for TSP. We now determine whether and how current architectures handle such information. 
To this end, we compare in Fig~\ref{encoder_issues_fig} the behavior of a  \textit{blind} model, from which edge weights are deliberately hidden, and the state-of-the-art version, which we name  \textit{edge-aware} here.  

Quite surprisingly, both the optimality gap and the absolute difference between rewards indicate that the blind counterpart is achieving better performance on most occasions: \emph{overall, this suggests  that  design-choices in the current state of the art  cannot deal  with edge features effectively}. Even worse, edge weights seem to be a source of noise that limits the agent's ability to learn the simplest policies. As an example of a simple policy, we noticed that using our reward function, solutions with mostly  minimum power levels performed better than random assignments. 

We further confirm this shortcoming by performing a similar comparison on the TSP. We evaluate a TSP version without coordinates as node features, but with distances between cities as edge features. Intuitively, removing node coordinates from the input should force the model to use edge weights instead to solve the problem. We compare, in Fig.~\ref{distance_encoding_back_tsp} 
(ignore distance-encoder for now)
, such \textit{edge-aware} version to a \textit{blind} TSP model and a model that uses coordinates. 
We observe that edge-aware architecture does not zero the gap with the one based on coordinates. Even though they perform better than  blind models, their representations appear to be less useful than when coordinates are available. This confirms that the encoding behaviour of this architecture is tainted by a 
poor handling of edge features, hinting to the need for alternative solutions.

\subsection{The decoder does not help} \label{decoder_issue_sec}

\begin{figure}[t]
\centering
\includegraphics[width=.9\linewidth]{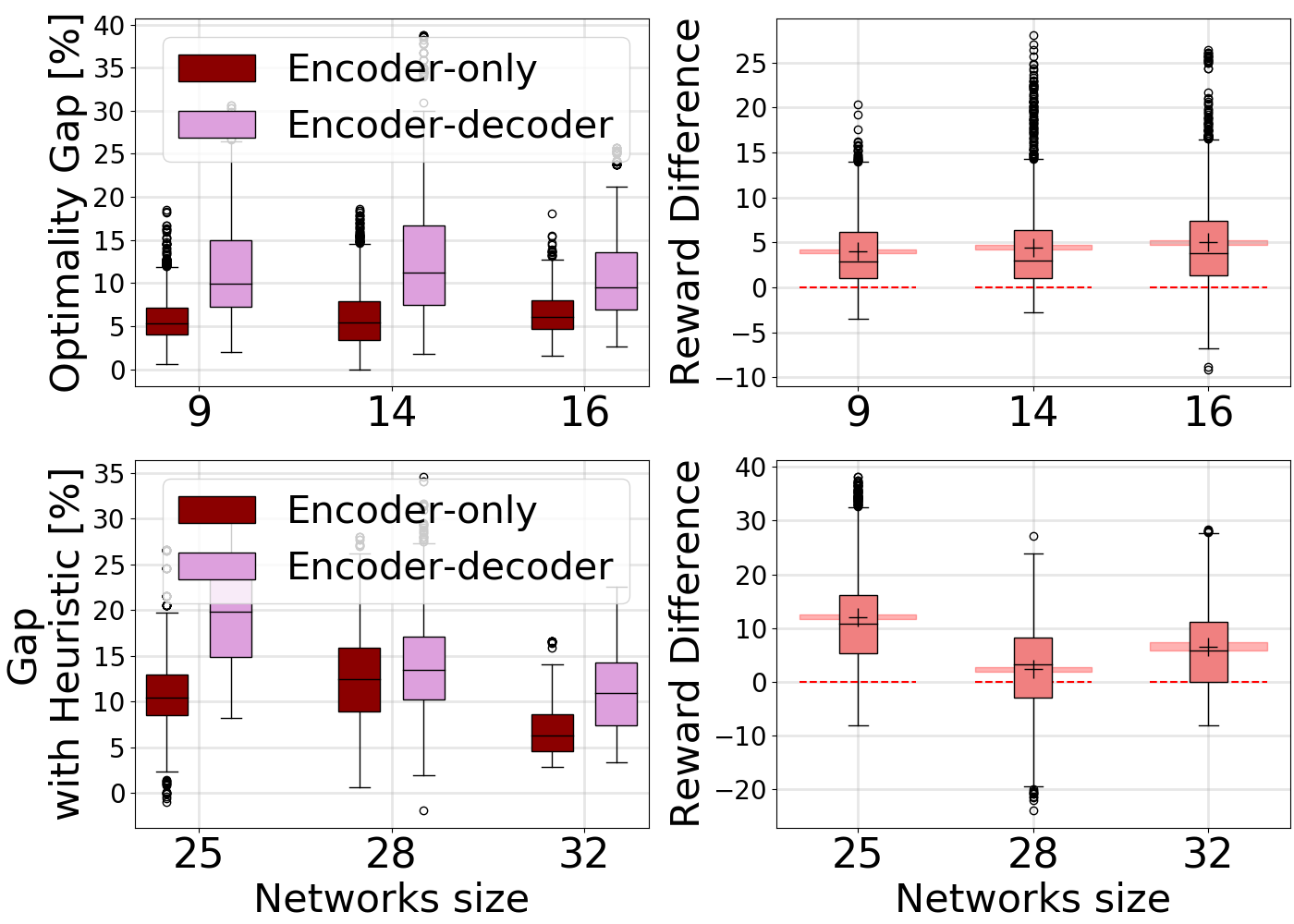}
\caption{Encoder-only vs Encoder-decoder}
\label{decoder_issue_fig}
\end{figure}

The other possible issue we discussed in Sec.~\ref{static_vs_dynamic} and Fig.\ref{decoder_issue} relates to static and action-dependent features. Since PCAP mainly has action-dependent features, it may  prove particularly challenging for current decoding schemes. The latter expect a context that is consistent through the decoding steps but obtain one that is continuously altered by intermediate decisions. To further investigate this, we test the same architecture ablating the decoder. 
Fig~\ref{decoder_issue_fig} compares an encoder-only model to the state-of-the-art encoder-decoder architecture. In the ablated architecture, the decoder context is removed and only a simple manipulation of the encoder's embeddings (e.g. linear transformation) is performed.


The results show that removing the decoder benefits all models, confirming our intuition that the decoder is probably confused, adding noise rather than benefiting the architecture. 
Not shown in the figure, the encoder-only model is the first attempt so-far that outperforms the \emph{100-random-attempts} baseline for the PCAP in four out of six scenarios. 
Overall, this result calls for better decoders that can handle  problems with action-dependent features.





\section{Distance Encoding to the rescue}\label{sec:rescue}
As introduced in Sec.~\ref{sec:related}, Distance Encoding has been proposed recently as a way to enhance the representational abilities of GNNs. 
After briefly introducing distance encoding, we empirically evaluate its impact on solving PCAP.

\subsection{Qualitative analysis}
Since existing models are unable to capture the relationships between nodes contained in graph edges, \emph{distance encoding} attempts to transfer edge information to nodes directly. Using a  ``edge2node transformation'', a novel artificial feature is added to node features which GNNs are more readily able  to process. In practice, the algorithm (i) obtains a distance matrix $D$ exploiting a certain definition of ``distance'' (e.g. shortest path, personalized PageRank score etc) (ii) apply some learnable transformations on the matrix (e.g. a simple dense NN layer) (iii) aggregates the information such that the $N \times N$ distance matrix $D$ becomes an $N \times 1$ vector. As a result, the latter is not only representative of the node itself, but also of its surroundings, allowing the use of all the available graph information and strengthening the structural representation of the problem.

Note that distance encoding creates a novel action-independent feature, making it possible to leverage the work of \cite{kool2018attention}. It becomes in theory possible to first use this new static information to build the encoded representation of the problem; then, with fixed and action-independent embeddings, add the dynamic information as a context, solving some of the decoder's shortcomings mentioned in Sec.~\ref{static_vs_dynamic}.


\begin{figure}[t]
\centering
\includegraphics[width=.5\linewidth]{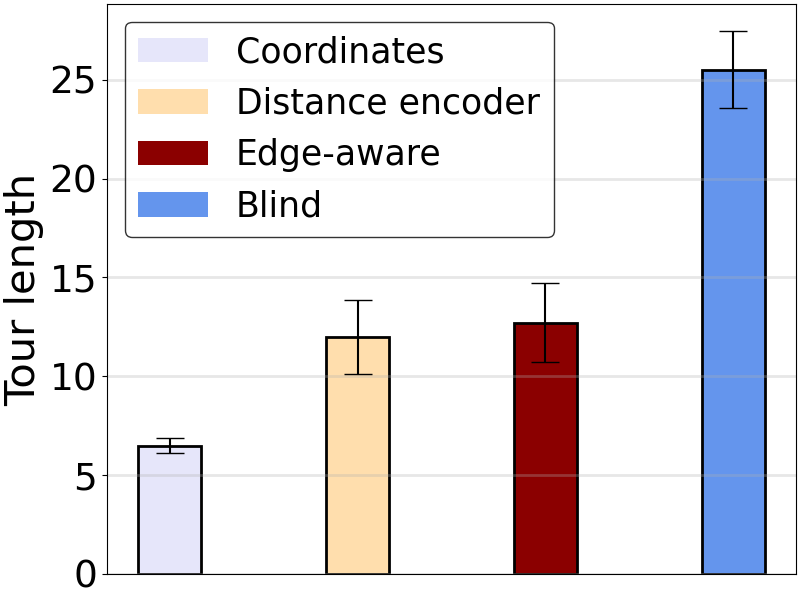}
\caption{Confirmation on TSP}
\label{distance_encoding_back_tsp}
\end{figure}



\subsection{Quantitative analysis}\label{de_to_the_rescue}
\begin{figure}[t]
\centering
\includegraphics[width=.9\linewidth]{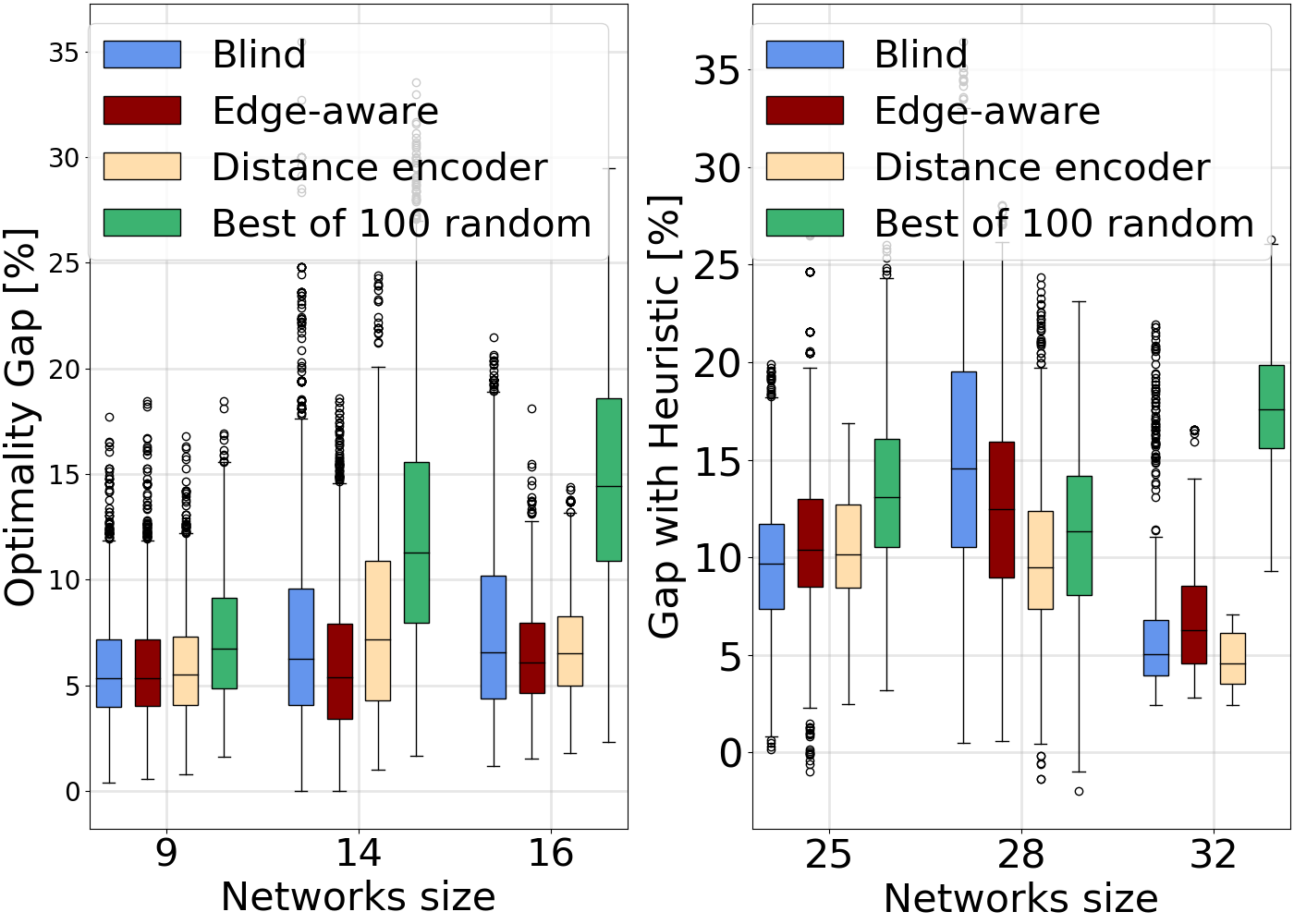}
\caption{Distance Encoding and PCAP}
\label{distance_encoding_pap}
\end{figure}

Fig.~\ref{distance_encoding_pap} shows our results when comparing a blind model, an edge-aware one, the usual random-based baseline and finally a model using Distance Encoding, first on the power allocation problem.
We observe that distance encoding  reduces the results' variance in almost all scenarios. 
The benefits of distance encoding are even more visible when considering larger and more challenging scenarios such as networks with 32 nodes.
Among the alternatives that we tested, distance encoding seems to consistently enhance the PCAP representation, as it is the only one that always outperforms the best choices of the random decision maker: e.g., notice especially for network size 28 that blind and edge-aware are outperformed by best of random choices.  

Similar observations concerning Distance Encoding are also valid for the TSP problem shown in  Fig.~\ref{distance_encoding_back_tsp}.  Although not spectacular, a significant improvement is brought by the Distance Encoding compared to the classic GNN counterparts  (blind  and edge-aware). 
At the same time, the fact that the distance encoding still under-performs compared to the coordinates version suggests clearly that, although it is a promising direction, more effort is needed to improve current encoders to successfully tackle problems where edge-features are of primary importance.

\section{Conclusions}
This paper evaluates whether state-of-the-art neural combinatorial optimization architectures, often evaluated on the TSP problem and similar variants, are useful for other graph problems of practical relevance, taking the power and channel radio allocation problem as a concrete example. Unfortunately, our results show that finding neural models capable of extracting good features from any input graph is still an unsolved problem: existing options we evaluated fail to properly exploit edge features. 
We also showed that the absence of immutable node features (such as coordinates for TSP) prevents existing decoders from leveraging a graph representation context that is coherent through the steps of the decoding loop: our experiments testify indeed that models using only the encoder's representations perform better than more sophisticated encoder-decoder counterparts. 

To enhance the encoder's representation power, we evaluate distance encoding, a recent proposal that transforms edge features into node ones. On the one hand, we find  that both our PCAP and a TSP variant without coordinates benefit from its use.  On the other hand, we also find that distance encoding does not attain the performance of coordinates-based counterpart. 
Ultimately, if building generic solvers for a variety of graph problems is the goal, our results call for carefully rethinking the neural combinatorial optimization pipeline along two axes: (i) encoders that better represent the inputs, and (ii) new ways to handle action-dependent features, which we believe to be interesting and important problems for the scientific community.

\bibliography{aaai22}

\end{document}